# Comparative Study on Supervised Learning Methods for Identifying Phytoplankton Species

Thi-Thu-Hong PHAN[1,2]

[1]Department t of Computer Science

Vietnam National University of Agriculture

Hanoi, Vietnam

Emilie POISSON CAILLAULT[2], André BIGAND[2]

[2]LISIC

Univ. Littoral Côte d'Opale

F-62228 Calais, France

*Abstract*— **Phytoplankton plays an important role in marine ecosystem. It is defined as a biological factor to assess marine quality. The identification of phytoplankton species has a high potential for monitoring environmental, climate changes and for evaluating water quality. However, phytoplankton species identification is not an easy task owing to their variability and ambiguity due to thousands of micro and pico-plankton species. Therefore, the aim of this paper is to build a framework for identifying phytoplankton species and to perform a comparison on different features types and classifiers. We propose a new features type extracted from raw signals of phytoplankton species. We then analyze the performance of various classifiers on the proposed features type as well as two other features types for finding the robust one. Through experiments, it is found that Random Forest using the proposed features gives the best classification results with average accuracy up to 98.24%.**

## I. INTRODUCTION

Phytoplankton is an important factor in environmental, economic and ecological policies. Being main producer of oxygen, phytoplankton is also an important food item in both aquaculture as well as mariculture. A question is raising: "how do changes in the global environment affect abundance, diversity, and production of plankton and nekton?" [1]. Many researchers show that environment changes strongly affect to phytoplankton and that it responds promptly to chemical perturbation [2-7]. The identification/classification of microscopic phytoplankton is therefore crucial for a wide variety of environmental monitoring applications in different domains such as: ecology (biodiversity), climate and economy. It is thus necessary to have a technique/tool capable of providing detailed description of phytoplankton species population from water samples.

Up to now, studies in identification/classification of phytoplankton species are usually carried out by visual comparing the collected profiles with references ones, or by the microscope method [8-9]. Using this microscope analysis method takes 3 to 4 hours for each sample (low frequency). It is laborious and extremely time-consuming. Hence, developing an automatic computer-aided machine system to identify/classify phytoplankton species is a required task.

Flow cytometry (FCM) analysis is a well-known and proven tool in aquatic ecology to quickly detect and quantify phytoplankton and bacteria (microorganism) from water samples [10-12]. "The various light scatter, diffraction, and fluorescence parameters measured by analytic FCM can provide characteristic "signatures" for each microbial cell, which allow taxa to be discriminated with the use of pattern-recognition techniques" [13]. Thus, the task of identifying phytoplankton species becomes the classification of multidimensional signals [14].

Regarding pattern-recognition techniques, a number of successful approaches have been proposed for automated identifying/ classifying of plankton species.

Concerning zooplankton identification/classification, several techniques including object classification technique for analyzing plankton images were developed by Hu and Davis [15] and Davis *et al.* [16]. In these two works, the images were collected from a video plankton recorder. A Support Vector Machine (SVM) is used for classifying a big image set (20,000 plankton images); the accuracy of classification on seven classes was achieved 71%. The performance of six classifiers: Multi-Layer Perceptron (MLP), K-Nearest Neighbors (5-NN), SVM (using linear and Radial Basis Function (RBF) kernels), Random Forest (RF), and C4.5 Decision Trees (DTs) were studied for classifying zooplankton images obtained the ZooScan system [27]. In this study, RF demonstrates the best performance and followed by SVM using the linear kernel. Irigoien *et al.* [28] carried out a research on classifying in zooplankton images with 17 categories and RF gives the highest result. The ZooScan digital imaging system for automatic analysis of zooplankton images is built by Grosjean *et al.* [1]. They tested individual classification algorithms as well as combinations of two or more different algorithms such as: double bagging associated with linear discriminant analysis, *K*-NN with discriminant vector forest and specifically mix of linear discriminant analysis with learning vector quantization, and random forest. Accuracy of the last combination achieves around 75% in the task of categorizing 29 zooplankton species. In the work of classifying binary zooplankton images, Luo *et al.* [17] investigated the performance of some classifiers, namely: SVM, RF, C4.5 DTs, and the cascade correlation neural network. SVM proves the highest classification performance with 90% and 75% on the six and seven classes, respectively.

Concerning phytoplankton species identification/ classification, many classification algorithms were used for this task such as Artificial Neural Networks (ANNs) using

FCM data [18-23] (72 phytoplankton species have been identified successful by ANN [20]). In another work, several methods namely: DTs, Naive Bayes (NB), ridge Linear Regression (LR), *K*-NN, SVM, bagged and boosted ensembles were applied to categorize phytoplankton images with 12 classes and an unknown class [24]. A system using SVM classifier for automated taxonomic classification of phytoplankton sampled with imaging-in-flow-cytometry is developed by Sosik and Olson[25]. In the work of Blaschko *et al.* [29], the accuracy of two modelling approaches for predicting boreal lake phytoplankton assemblages and their ability to detect human impact were studied. They used random forest to predict biological group membership and species. Verikas *et al.* [26] have recently investigated to detect, recognize, and estimate abundance objects representing the *P.minimum* species in phytoplankton images. The classification performance of SVM and RF methods was compared on 158 phytoplankton images.

It is found that the number of studies using plankton signals (FCM data) is less than the ones using plankton images. Most of studies based on signals used available features generated from a FCM system. However, only a few earlier studies used FCM signals (both available features and raw signals) to compare the performance of classification methods [14]. In addition, RF has proved its performance in many applications of plankton species identification/classification [1, 27, 28, and 29]. With the best of our knowledge, there is no application that combines the FCM signals and RF to determine phytoplankton species.

Therefore, our main contributions in this paper are: (1) propose a new features type extracted from the raw signals of phytoplankton species; (2) perform a comparative analysis of identifying phytoplankton species using a variety of advance machine learning models such as *K*-NN (1-NN), SVM, RF and several modification versions of RF. This permits one to determine the best features for representing phytoplankton species and classifier for classifying phytoplankton species with high accuracy. The paper is organized as follows. Section 2 introduces materials and methods. Section 3 demonstrates our experimental results and discussion. Conclusion and future works are drawn in Section 4.

## II. MATERIALS AND METHODS

### A) Data presentation

In this study, we reuse the data of our previous study [14] (Data presentation and Signal acquisition).

The data is acquired from 7 culture samples, whose particles belong to 7 distinct phytoplankton species: *Chaetoceros socialis, Emiliania Huxleyi, Lauderia annulata, Leptocylindrus minimus, Phaeocystis globosa, Skeletonema costatum* and *Thalassiosira rotula*. Each species is equally represented by 100 shape-profiles and each culture sample was labeled by biologists using a microscope [9]. So, the data set has 700 (100×7) phytoplankton cells.

### B) Signal acquisition

Multi-signals were gathered in the LOG laboratory[1] from different phytoplankton species living in Eastern Channel, with a CytoSense flow cytometer (CytoBuoy[2]), and labeled by biologists [9] once having them isolated from the natural environment. Flow cytometry is a technique used to characterize individual particles (cells or bacteria) derived by a liquid flow at high speed in front of a laser light. Different signals either optical or physical are provided: forward scatter (reflecting the particle length), sideward scatter (being more dependent on the particle internal structure) and several wavelengths of fluorescence that depend upon the type of its photosynthetic pigments measures.

More precisely, in the used signals library, each detected particle is described by 8 mono-dimensional raw signals issued from the flow cytometer in identical experimental conditions (same sampling rates, same detection threshold, etc.):

- one signal on forward scatter (FWS), corresponding to the cell length;

- two signals on sideward scatter (SWS), corresponding to the internal structure, in high and low sensitivity levels (SWS HS, SWS LS);

- two signals on red fluorescence (FLR), $\lambda em > 620nm$, in high and low sensitivity (FLR HS, FLR LS), which characterize chlorophyll pigments;

- one signal on orange fluorescence (FLO), $565nm < \lambda em < 592nm$, in low sensitivity (FLO LS);

- two signals on yellow fluorescence (FLY), $545nm < \lambda em < 570nm$, in high and low sensitivity (FLY HS, FLY LS).

These signals are composed of voltage measures (mV), and their sampling period was here chosen to correspond to 0.5μ-meter displacement of the water flow. Consequently, the longer the cell is, the higher the number of sampled measures is, and the time axis can be interpreted as a spatial length axis. Phytoplankton species identification is a hard task so all these signals are used to make the particles characterization. Each particle of our experiment is consequently characterized by the 8 signals described above. Figures in Fig 1. show some signal samples *of Lauderia annulata* and *Emiliania huxleyi* species.

---

[1] Laboratoire d'Oceanologie et de Géosciences, UMR 8187: http://log.univ-littoral.fr

[2] Cytobuoy system: http://www.cytobuoy.com

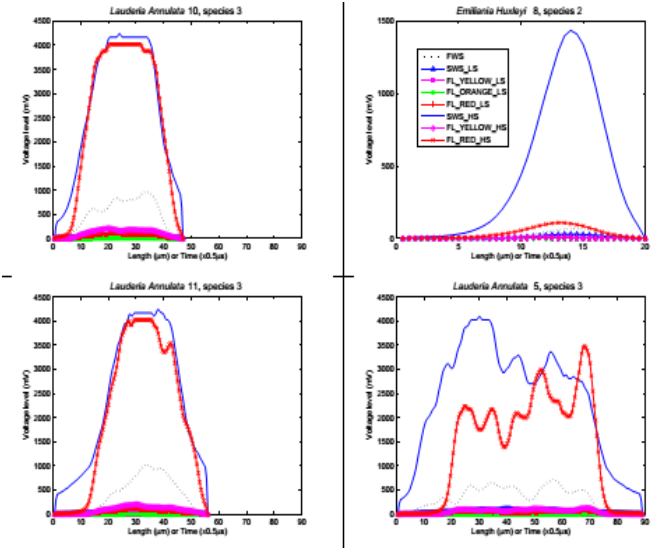

Fig. 1. 8D-signals describing two species

*C) Phytoplankton descriptor*

After acquiring raw signals of a phytoplankton from the FCM system, phytoplankton descriptor must be computed to represent the phytoplankton species, which will be presented to a classifier. The phytoplankton descriptor describes properties of a phytoplankton cell (for example length, number of peaks… of each raw signal or the ratio of dissimilarity of each pairs of phytoplankton cells). In this paper, these properties are typically called "features". We investigate three features types: derived features, proposed features and dissimilarity features [14].

1. *Derived features*

For each signal, 4 elements are extracted by a Cytobuoy machine including: length, height, integral, and number of peaks. So each phytoplankton cell is presented by a vector of 32 features.

2. *Proposed features*

The idea of our proposal is to offer some features that can better represent dynamics and shape of phytoplankton signals. Among of the possible features, signal moments and entropy give better results. Denoted $Q=\{q_1, q_2... q_n\}$ are the values of each phytoplankton signal (curve). With each raw signal, 9 elements are calculated as following:

- Percentile: The $m^{th}$ percentile of Q's values is the value that cuts off the first m percent of Q's values when these values are sorted in ascending order (m=30 is used in this study).

- Max: It is the maximum of Q's values:

$$\max(Q) = max\{q_1, q_2, ..., q_n\}$$

- First moment: It is the mean of Q's values:

$$\bar{Q} = mean(Q) = \frac{\sum_{j=1}^{n} q_j}{n}$$

- Standard derivation: It is the standard derivation of Q's values, based on the 2nd moment central:

$$std(Q) = \sqrt{\frac{1}{n}\sum_{i=1}^{n}(q_i - \bar{Q})^2} = \sqrt{\mu_2}$$

- Median: The median of Q's values is the value separating the higher half and the lower half. It is the middle number when the data is sorted from lowest value to highest value.

- Third moment: It is the 3rd moment of Q's values:

$$\gamma_1 = \frac{\mu_3}{\mu_2^{3/2}}$$

where $\mu_2$ and $\mu_3$ are the second and third central moments. $\gamma_1$ is the normalized 3$^{rd}$ moment central (Skewness coefficient). We can know the data distribution thanks to this coefficient. In this study, we use the 3$^{rd}$ moment instead of utilizing the Skewness coefficient because some signals have all 0 values.

- Nop: It is the number of peaks of Q's values calculating based on the second derivative.

- Length: It is the length of the curve.

- Entropy: It is based on the Shannon entropy formula:

$$entropy(Q) = -\sum_{i=1}^{n} p_i log(p_i), \text{ with } p_i = \frac{q_i}{\sum_{i=1}^{n} q_i}$$

Consequently, the proposed features vector is 72 dimensions.

3. *Dissimilarity features*

Dynamic Time Warping (DTW) [30] is an algorithm to align two sequences (may vary in time) by warping the time axis until finding an optimal matching between the two sequences according to suitable metrics. However, it is not easy to interpret the cost matching. Thus, Caillaut et al. [14] proposed a dissimilarity distance that adapts the DTW matching cost and can deal with multidimensional signals. They replaced the distance DTW d (L1–distance or L2–distance) with a DTW dissimilarity $s$ ($s \in [0, 1]$-normalized dissimilarity degree):

$$s(q_{i_k}, r_{j_k}) = \frac{d(q_{i_k}, r_{j_k})}{\max\{d(q_{i_k}, 0), d(r_{j_k}, 0)\}}$$

In which $Q=\{q_1, q_2,...,q_n\}$ and $R=\{r_1, r_2,...,r_m\}$ are the two signals of different size. The algorithm makes a matching $P = \{(i_k, j_k), k = 1... n_k, i_k = 1... n, j_k = 1... m\}$ between the points of Q and R signals, according to some time conditions.

Therefore, each phytoplankton cell is presented by a vector of 700 dissimilarity features, in which a feature is the DTW dissimilarity between this cell and one cell in the data set.

*D) Classification*

After feature extraction, a classifier is learned for identification of different phytoplankton species. In the following, we review some prominent classification models:

1. *K-Nearest Neighbors*

*K*-nearest neighbors [32] has been widely used in classification problems because it is simple, effective and non-parametric [31]. For each sample of a test set, we found *K* cases in the train set that is minimum distance between the feature vectors of the sample and those of the train set. A decision of the label of a new sample is based on majority vote of the $K$ label found.

### 2. Support Vector Machine

The basic idea of support vector machine [33] is to find an optimal hyper-plane for linearly separable patterns in a high dimensional space where features are mapped onto. The work is to detect the one that maximizes the margin around the separating hyper-plane from training set. A decision of the label of a new sample is based on its distance with the trained support vectors.

### 3. Random Forest

Breiman [34] proposed random forest, a classification technique by constructing an ensemble of decision trees. In which each decision tree uses a different bootstrap sample of the response variables and at each node, a small subset of randomly selected variables from original ones for the binary splitting. For predicting new data, a RF aggregated the outputs of all trees.

### 4. Regularized Random Forest (RRF), Guided RRF (GRRF), Guided RF (GRF)

RRF, GRRF, GRF are different modification versions of the original RF. But these methods are just similar to initial RF method in the step of predicting new data, and they are different in step of finding features to build each decision tree of forest. Indeed, RRF was proposed for improving feature selection on the decision tree by limiting the choice of new feature at each tree node and evaluating features (using Gini index) on a part of the training data [35]. This process of feature selection is greedy because variables are selected based on a subsample of data variables at each node.

GRRF [36] is an enhanced RRF. This approach uses the feature importance scores generated from an initial random forest to guide the feature selection process in RRF for avoiding of selecting not strongly relevant features. While GRRF selects a subset of relevant and non-redundant features, GRF selects a subset of "relevant" features. So GRF often selects a lot more features than GRRF (sometimes most of the features), but it may lead to better classification accuracy than GRRF. Nevertheless, each tree of GRF is built independently and GRF can be implemented in a distributed computing framework [37].

## III. EXPERIMENT AND DISCUSSION

We have conducted a set of experiments on various features types and classification models to evaluate their performance on phytoplankton species data (as mentioned above).

### Experiment set up

To conduct all experiments, we use a computer with 64 bits Window 7, core i7, CPU 3.0 GHz and 8 GB main memory. For computing proposed features we use the following R-packages: base, stats, moment [38], and entropy [39]. We utilize the latest R-packages of RF [34], RRF (RRF, GRRF, and GRF) [37], e1071 package (SVM) [40], class package ($K$-NN) [41] for classifying. Other R-packages like FactoMineR [44], lda [42], have been used to find the most important features.

Concerning SVM, after testing different kernels on different features types, we choose polynomial kernel of SVM (degree =3) for the derived features and the dissimilarity features, RBF kernel of SVM for the proposed features (tune.svm function [40] is used to find out the optimal coefficients ($\gamma$=0.01 and $C$=32, for example). With $K$-NN, one of the most important parameters is to choice of suitable value of $K$. In our experiment, we test with different values of $K$ ($K$ = 1 to 10) and this model gives the best results when $K$ = 1. For RF, the basic two parameters are specified to train the model are: $ntree$ - number of trees to be constructed in the forest and $mtry$ - number of input variables randomly sampled as candidates at each node. In this study, $ntree$=500 is fixed for all RF versions. $mtry$ of RF, $\gamma$ of GRRF and $\gamma$ of GRF are default values: the square root of the number of features [34], 0.1 [36] and 1 [37], respectively.

Each classifier is evaluated using a 4-fold cross validation to determine the recognition error rate and this cross validation is repeated 10 times. The data set of 700 (100×7) phytoplankton cells is divided into 4 subsets of 175 (25 × 7) cells. Each subset respects an equal target distribution. The learning phase uses three subsets and predicts the remains as test set. For classifying phytoplankton species, in the first step, we extract proposed features (derived features are available) and calculate dissimilarity of each pairs phytoplankton cells from the raw signals. In the next step, after finishing of the learning process, the classification models are used to predict test set. The accuracy average of classification methods are given in Table I, II, III. The results of contingency table between different models and between different features types of one in the 10 iterations are presented in Table IV, V.

TABLE I. ACCURACY OF TEST RECOGNITION OF DIFFERENT CLASSIFICATION MODELS ON THE DERIVED FEATURES (%)

| Classifier | SVM | $K$-NN | RF | RRF | GRRF | GRF |
|---|---|---|---|---|---|---|
| Fold1 | 95.89 | 88.63 | 96.91 | 95.37 | 95.66 | 95.94 |
| Fold2 | 94.06 | 86.80 | 96.17 | 95.26 | 96.06 | 95.43 |
| Fold3 | 95.03 | 87.60 | 96.63 | 95.54 | 96.06 | 94.97 |
| Fold4 | 94.63 | 88.57 | 96.86 | 96.23 | 96.46 | 95.37 |
| Average | 94.90 | 87.90 | **96.64** | 95.60 | 96.06 | 95.43 |

TABLE II. ACCURACY OF TEST RECOGNITION OF DIFFERENT CLASSIFICATION MODELS ON THE PROPOSED FEATURES (%)

| Classifier | SVM | $K$-NN | RF | RRF | GRRF | GRF |
|---|---|---|---|---|---|---|
| Fold1 | 96.74 | 82.11 | 97.65 | 95.89 | 95.77 | 96.86 |
| Fold2 | 97.54 | 83.12 | 98.57 | 96.29 | 94.83 | 97.37 |
| Fold3 | 97.66 | 82.97 | 98.63 | 96.97 | 96.86 | 97.26 |
| Fold4 | 97.32 | 82.74 | 98.12 | 96.68 | 96.69 | 97.54 |
| Average | 97.31 | 82.74 | **98.24** | 96.46 | 96.03 | 97.26 |

TABLE III. ACCURACY OF TEST RECOGNITION OF DIFFERENT CLASSIFICATION MODELS ON THE DISSIMILARITY FEATURES (%)

| Classifier | SVM | $K$-NN | RF | RRF | GRRF | GRF |
|---|---|---|---|---|---|---|
| Fold1 | 94.29 | 97.31 | 97.66 | 94.74 | 94.74 | 96.40 |
| Fold2 | 94.91 | 97.72 | 97.43 | 95.66 | 94.34 | 96.57 |
| Fold3 | 94.86 | 97.54 | 97.20 | 94.29 | 93.83 | 95.77 |
| Fold4 | 94.97 | 97.20 | 97.49 | 95.54 | 94.57 | 96.57 |
| Average | 94.76 | **97.44** | **97.44** | 95.06 | 94.37 | 96.33 |

The reliability of classification models is evaluated based on classification accuracy of the test sets. The classification results of six methods using different features types are illustrated in Tab.I, II, III. These tables show that RF has the highest classification accuracy on all types of features when comparing to other classification methods. RRF, GRRF and GRF are improved versions of RF but they are recommended for high-dimensional data. In this study, all features types are not high-dimensional (number of dimensions are 32, 72, and 700 respectively), the RRF, GRRF as well as GRF therefore do not give the best results but they also provide good results on all features types.

Table I and Table II present the results of different classification models on the derived features and the proposed features. Regarding these two kinds of features, RF has proven the best capability for classifying on all folds, with classification accuracy average 96.64% (Tab. I) and 98.24% (Tab. II). The *K*-NN model and SVM model show a lower classification rate compared to all versions of RF with 87.90% and 94.90%, respectively (Tab. I).

Table II shows that when combining SVM with proposal features gives better results (97.31%) than combining SVM with derived features (94.9%, Tab. I) and dissimilarity features (94.76, Tab. III). In contrast to the SVM, *K*-NN has the lowest performance (82.74%, Tab. II), which implies that combining *K*-NN with the proposed features as well as with the derived features is not favor for identifying phytoplankton species. This method drops its performance (Tab. II) because it is very sensitive to the $3^{rd}$ moment (the $3^{rd}$ moment domain is from 0 to 69,000,000 while other features domain is too small). Besides, for more robust verification of the proposal features and classifiers, 5-fold cross validation is performed, in which 3 folds for learning, 1 fold for validation and 1 fold for testing. RF method always proves the best performance 98.57%, following by GRF 97.86%. SVM and GRRF have the same accuracy 97.14%. The performance of GRF is 95.71% and the last is *K*-NN with 79.29%.

Table III illustrates the classification results of different methods on the dissimilarity features. In contrast to the results in Tab. I and Tab. II, K-NN method demonstrates superior capability in task of identifying phytoplankton species. This result is entirely interpretable because through experimental tests prove that the combination of 1-NN with DTW distance "has proven exceptionally difficult to beat" [43]. Concerning RRF, GRRF and GRF, with this type features, the performance of these methods are less than their performance when they combine with the derived features and with the proposed features. However, RF has always stable in the best classification capacity, the same result as *K*-NN 97.44%.

In addition, in this paper we also compare the results of target assignment of the same classifier on different features types (Tab. IV) as well as different classifiers on the same features type (Tab. V). Table IV is a contingency table of RF classifier on the derived features and the proposed features. In the $1^{st}$ fold, RF classifies correctly 165 samples on the proposed features and 171 samples on the derived features. However, only 164 samples are the same classified on the both of features types. Table V is a contingency table of *K*-NN and RF methods based on the dissimilarity features. In the $4^{th}$ fold, both RF and *K*-NN methods correctly classify 169 samples but only 167 samples are classified in common.

Also the comparison of performance of different classifiers and results of target assignment, we carry out identifying which attribute affects the response variable (true label) on the derived features and the proposed features. A supervised technique: Linear Discriminant Analysis (LDA) [42] is used for analyzing. This technique permits to detect a linear combination of predictor variables (features) that best characterizes or separates two or more classes (targets). In fact, with the derived features, the hflo_ls feature (the height of signal on orange fluorescence FLO in low sensitivity, which corresponds to the *maximum* feature of the proposed features) is strong relative to the target variable (28.29% of contribution for all LD components). With the proposed features: the entropy_flo_ls variable (the entropy of signal on orange fluorescence FLO in low sensitivity) is the most important feature which affects the classification variable (46.15%). This result shows that, on the 8 signals, the signal on orange fluorescence FLO in low sensitivity is the most influential to the response variable. On the other hand, the classification results of all RF versions using the proposed features (Tab. II) are higher than their results using the derived features (Tab. I). From these analyses, we find that the proposed features are very significant for the task of classifying phytoplankton species.

Based on the results of classification of seven phytoplankton species (Tab. I, II, III), RF has proven its ability and stability for identifying phytoplankton species as combining with different features types (the best performance when RF combining with the proposed features of 98.24%). In contrast, SVM and *K*-NN indicate less classification capability on the derived features and the proposed features although different kernels have been used and the parameters have been optimized to achieve the best result.

TABLE IV. CONTINGENCY TABLE OF RF MODEL ON THE DISSIMILARITY FEATURES AND THE PROPOSED FEATURES (T: TRUE LABEL, F: FALSE LABEL)

| **Random Forest** | | **Proposed features** | | | | | | | |
|---|---|---|---|---|---|---|---|---|---|
| | | *Fold 1* | | *Fold 2* | | *Fold 3* | | *Fold 4* | |
| | | *T* | *F* | *T* | *F* | *T* | *F* | *T* | *F* |
| Derived features | T | 164 | 7 | 165 | 2 | 166 | 1 | 168 | 3 |
| | F | 1 | 3 | 8 | 0 | 0 | 9 | 3 | 1 |

TABLE V. CONTINGENCY TABLE OF RF AND K-NN MODELS ON THE DISSIMILARITY FEATURES

| **Dissimilarity features** | | ***K*-NN** | | | | | | | |
|---|---|---|---|---|---|---|---|---|---|
| | | *Fold 1* | | *Fold 2* | | *Fold 3* | | *Fold 4* | |
| | | *T* | *F* | *T* | *F* | *T* | *F* | *T* | *F* |
| RF | T | 171 | 3 | 168 | 2 | 170 | 0 | 167 | 2 |
| | F | 0 | 1 | 2 | 3 | 3 | 2 | 2 | 4 |

RF has high accuracy classifier and stability because for the classification situation, Breiman [34] pointed out that accuracy of classification can be improved by aggregating the results of many simple classifiers that have little bias by averaging or voting.

From the above results and analysis, we suggest combining the proposed features with RF for identifying of phytoplankton species.

IV. CONCLUSION AND FUTURE WORKS

This paper proposes a quantitative comparison of performance of six classification methods for identifying phytoplankton species. The obtained results prove that RF with the proposed features is the best robust for phytoplankton species identification. The paper highlights two mains contributions. Firstly, we propose new features extracted from raw FCM signals. Secondly, we provide a quantitative comparison of different classification algorithms applied to different features types. Besides, we also compare target assignment of the same classifier on different features types as well as different classifiers on the same features type. In addition, we carry out analyzing on the derived features and the proposed feature to identify which attribute affect the target variable. The present work will permit combining classifiers (e.g. RF method with *K*-NN method) or features types (e.g. the derived features with the prosed features) to improve classification results.